\documentclass[letterpaper, 10 pt, conference]{ieeeconf}  

\IEEEoverridecommandlockouts                              
\usepackage{graphicx}
\usepackage{amsmath}
\usepackage{cite}
\usepackage{url}

\usepackage{times}
\usepackage{fancyhdr,graphicx,amsmath,amssymb}
\usepackage[ruled,vlined]{algorithm2e}

\usepackage[font=footnotesize,belowskip=-0.5pt,aboveskip=0pt]{caption}
\usepackage{bigstrut}
\usepackage[utf8]{inputenc}
\usepackage{array}
\usepackage{multirow}
\usepackage{tabularx}
\usepackage{bigstrut}
\usepackage{makecell}
\usepackage{tabu}
\usepackage{bm}
\usepackage{caption}

\usepackage{amsmath}          
\usepackage{graphics}
\usepackage{graphicx}
\usepackage{float}
\usepackage{caption}
\usepackage{subcaption}
\usepackage{dblfloatfix}
\usepackage{calc}
\usepackage{url}
\usepackage{textcomp}
\usepackage{balance}
\usepackage{siunitx}
\usepackage{mdwlist}
\usepackage{comment}

\usepackage{gensymb}

\usepackage{array}
\usepackage{multirow}
\usepackage{tabularx}
\usepackage{booktabs}

\usepackage{bigstrut}
\usepackage{fixltx2e}

\usepackage{color}
\usepackage{hhline}

\usepackage[capitalize]{cleveref}

\title{\LARGE \bf
PinFT: Miniature 5-Axis Force/Torque Sensor Embeddable to Tweezer-like Tool
}

\author{Leo King, Jenny Chen, Tae Myung Huh
\thanks{All authors are with Dept. of Electrical and Computer Engineering, University of California Santa Cruz, Santa Cruz, CA, USA. {\tt\small thuh@ucsc.edu}}%
}

\begin{document}

\maketitle

\thispagestyle{empty}
\pagestyle{empty}

\global\csname @topnum\endcsname 0
\global\csname @botnum\endcsname 0

\begin{abstract}
We present PinFT, a miniature five-axis capacitive force/torque sensor designed for direct tip-level integration into tweezer-like tools. 
The sensor employs a compact three-PCB stack with segmented plated through-hole electrodes and a silicone elastomer dielectric, enabling five-degree-of-freedom force and torque sensing ($F_x$, $F_y$, $F_z$, $T_x$, $T_y$) through displacement of a central 2\,mm-diameter stainless steel pin. 
The fabricated prototype was calibrated using a higher-order polynomial mapping, yielding mean absolute errors of approximately 0.23\,N for forces and 2.5\,mN$\cdot$m for torques, with coefficients of determination ($R^2$) exceeding 0.97 across all axes. 
To demonstrate practical utility, a 3D-printed tweezer integrating PinFT sensors at both tips was mounted on a parallel-jaw gripper and evaluated across three representative manipulation tasks: grasping a sub-millimeter SMD capacitor, pulling a simulated hair from a silicone substrate, and tearing a compliant silicone specimen. 
In all cases, per-tip force sensing reliably captured characteristic force signatures that distinguish successful manipulation from failure events---including slip and object ejection---using gradient-based features derived from internal grasp force and net interaction force. 
These results demonstrate that direct, per-tip force sensing enables standard parallel-jaw grippers to monitor and interpret fine manipulation tasks performed through a handheld tweezer.

\end{abstract}


\section{Introduction}
\label{sec:Intro}

Humans routinely use tools to extend their manipulation capabilities, achieving levels of precision and delicacy that are difficult to attain with bare hands alone. Instruments 
like tweezers enable reliable interaction with small fragile objects essential in domains ranging from microsurgery to electronics assembly. The effectiveness arises from complementary factors: mechanical leverage attenuates finger forces to enable controlled, low-force interaction at the tool tips, while slender contact geometries provide access and precision at size scales where 
fingertips are too large. Moreover, during tool use, humans perceive the tool as an extension of the body, with tool-mediated haptic signals conveying contact, force, and slip information. Localizing sensation at the tool-tip, effectively acting as an extended fingertip \cite{miller_sensing_2018}.

Robotic manipulation could benefit from similar principles by leveraging handheld tools, such as tweezers, to achieve fine and delicate interactions beyond the capabilities of conventional parallel-jaw grippers. Prior work has explored specialized end-effectors that emulate forceps- or tweezer-like kinematics\cite{verotti_comprehensive_2017}; however, broader usability would be achieved if standard robotic grippers could flexibly manipulate a variety of task-specific tools. A key challenge in enabling such tool-mediated manipulation is providing the robot with tactile or force feedback that accurately reflects interactions occurring at the tool–object interface.

\begin{figure} [tbp!]
    \centering
    \includegraphics[width=1\linewidth]{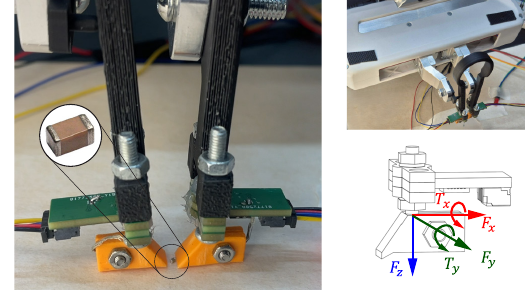}
    \vspace{0.5mm}
    \caption{The PinFT-integrated robotic tweezer system.
Left: Close-up of the dual-tip configuration, where each tweezer tip is instrumented with a miniature PinFT five-axis capacitive force/torque sensor. The inset shows a sub-millimeter SMD capacitor used in precision grasping experiments. Top right: Integration of the 3D-printed tweezer tool mounted on a parallel-jaw gripper. Bottom right: Definition of the local force and torque coordinate frame ($F_x$, $F_y$, $F_z$, $T_x$, $T_y$) measured at each tip through the central 2 mm stainless steel sensing pin.}
    \label{fig:CapBlkDiagram}
\end{figure} 


One approach is to infer tool–object interaction forces using tactile or force sensors mounted at the gripper \cite{li2025grasp}. Such methods require explicit modeling of tool geometry, compliance, and dynamics\cite{guo2019compensating} to estimate contact forces at the tip, often necessitating tool-environment-specific calibration\cite{nasiri2024admittance} which limits generalization across tools and tasks.

An alternative is to directly sense interaction forces at the tool itself. Existing force-sensing forceps and tweezers typically place sensors at the joining feature of the two tips, measuring the net force applied to an object once it is rigidly grasped\cite{gonenc20173,liu2023haptics}. However, tip-level sensing—measuring forces independently at each side of the tweezer—can provide richer and more localized information about contact onset, force distribution, and incipient slip at each jaw. While some surgical instruments integrate tip-level force sensors\cite{kim2015force,puangmali2008state}, these designs are tightly coupled to specific tool geometries and lack modularity, making adaptation to new tools or tip geometries cumbersome.

In this paper, we propose a miniature force–torque sensor that enables direct measurement of interaction forces at the tool tip, as shown in \cref{fig:CapBlkDiagram}. The main contributions are:
\begin{itemize}
    \item A compact capacitive force–torque sensor design capable of measuring five-axis force and torque using a simple three printed circuit board (PCB) stack. The resulting slender form factor enables seamless integration into a tweezer-like tool, while allowing customizable tool tips to be easily fabricated and mounted.
    \item Demonstration that direct tool-tip force sensing enables standard parallel-jaw grippers to monitor fine manipulation tasks performed through a handheld tweezer.
\end{itemize}

\subsection{Related Work: Tool-Mounted Force and Tactile Sensors}


Prior work has shown placing force or tactile sensors directly on a tool/tool's tip improves force measurement fidelity by capturing interaction forces at the point of contact, an approach widely adopted in medical/surgical robotics \cite{puangmali2008state}.

Distal sensing has been integrated into a range of surgical instruments. Tool-tip force sensors have been incorporated into surgical drills to isolate drill--tissue interaction forces from non-contact loads \cite{chen2023drill}, and into microsurgical instruments using fiber Bragg grating (FBG) sensors to resolve sub-millinewton forces \cite{he2014fbg}. Several studies embed sensing elements directly into forceps or gripper jaws, including strain-gauge-based micro-forceps for microsurgery \cite{menciassi2001forceps, hricko2011raad}, capacitive force-sensing surgical forceps \cite{kim2015force}, and motorized micro-forceps with integrated 3-DOF force sensing for vitreoretinal surgery \cite{gonenc20173}. More recently, modular jaw-mounted force sensors using miniature MEMS load cells have enabled multi-axis force measurement near the tool tip in robot-assisted minimally invasive surgery \cite{chua2023modular}, while multimodal, haptics-enabled forceps have been proposed to support task-autonomous surgical behaviors \cite{liu2023haptics}.

Despite these advances, existing tool-mounted sensors face notable limitations. Many rely on bulky commercial force/torque sensors \cite{chen2023drill} or specialized MEMS- and fiber-optic-based fabrication \cite{menciassi2001forceps, hricko2011raad, he2014fbg}, limiting accessibility and reproducibility. Moreover, most miniature designs measure only one to three force components \cite{gonenc20173, chua2023modular}, and sensing in forceps-like tools is often implemented at intermediate structural rather than at the individual tips. As a result, internal grasping forces cannot be directly measured, and secure grasp is typically assumed, yielding limited information about the grasp state. Existing tip-level sensing solutions are further highly customized for specific surgical instruments \cite{kim2015force, puangmali2008state}, restricting their applicability to a broader range of tool tips, including general-purpose tweezer-like tools.

\section{Design of PinFT}

\begin{figure}[t]
    \centering

    \begin{subfigure}[c]{0.5\columnwidth}
        \centering
        \includegraphics[width=\linewidth]{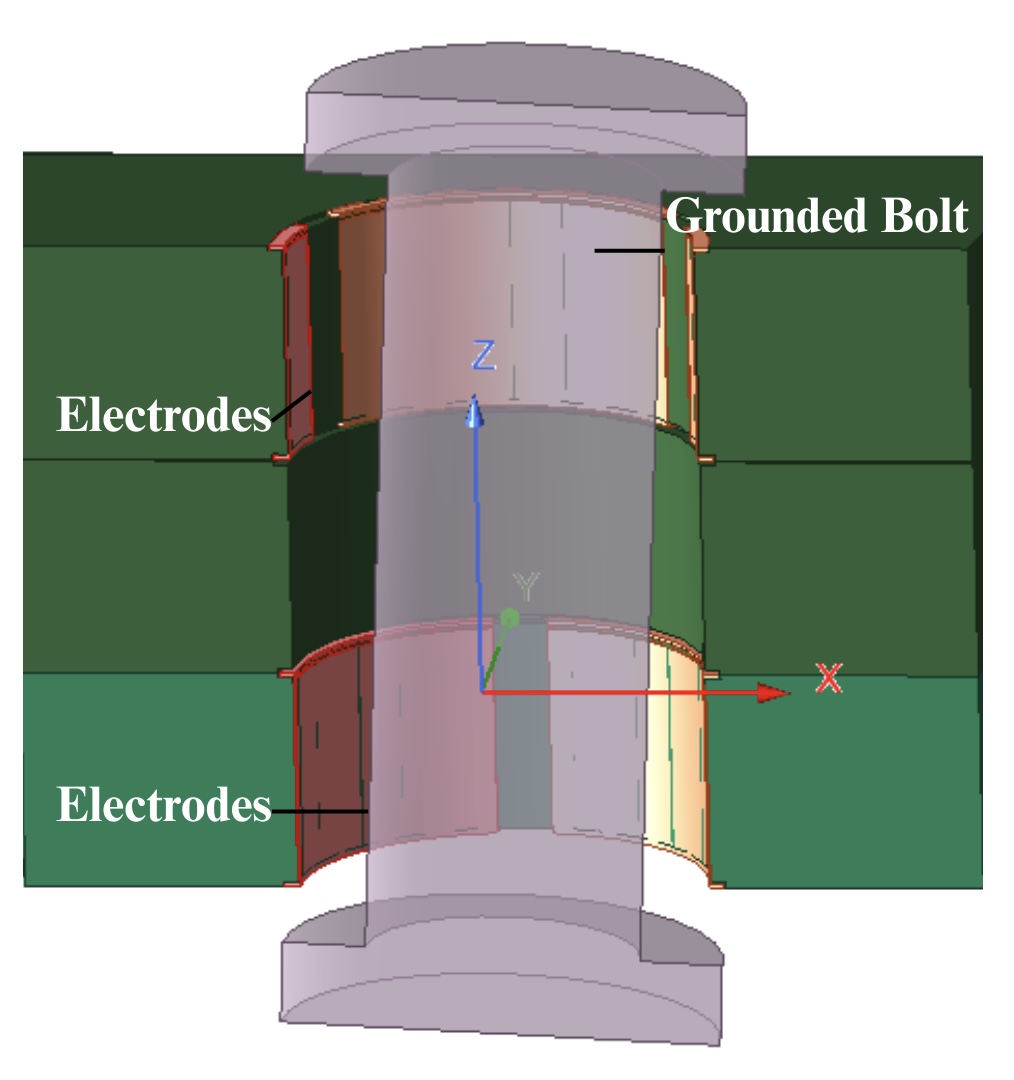}
        \label{fig:DesignRot}
    \end{subfigure}
    \hfill
    \begin{subfigure}[t]{0.30\columnwidth}
        \centering
        \includegraphics[width=\linewidth]{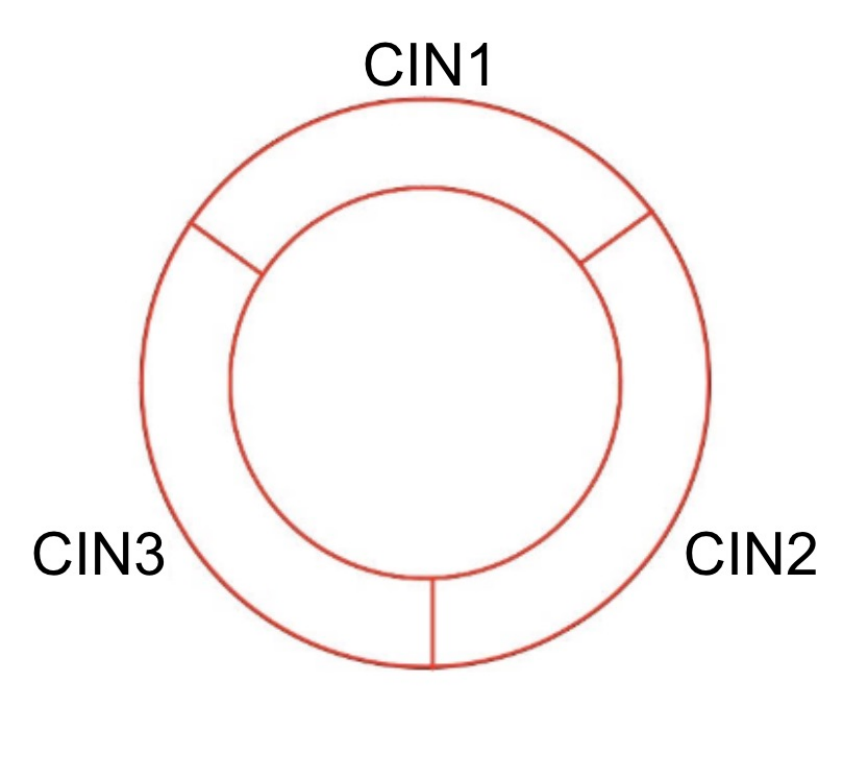}

        \vspace{6pt}

        \includegraphics[width=\linewidth]{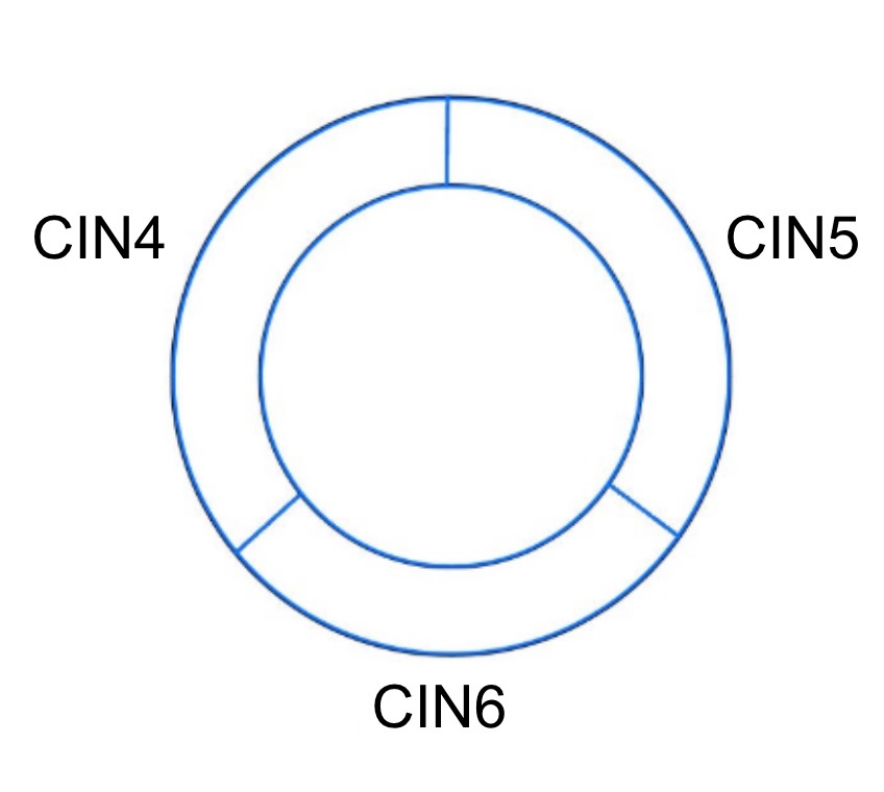}
    \end{subfigure}

    \caption{ Cross-sectional view of the PinFT sensor showing the grounded central pin surrounded by segmented plated through-hole electrodes (left), and the top (CIN1--CIN3) and bottom (CIN4--CIN6) PCB electrode layouts (right)}
    \label{fig:CombinedElectrodeViews}
\end{figure}

To enable force–torque sensing at the tip of tweezer-like hand tools, we propose a compact stacked PCB-based capacitive sensor with a silicone elastomer dielectric and segmented plated through-hole electrodes, as shown in \cref{fig:CombinedElectrodeViews}. This architecture achieves five-degree-of-freedom sensing—three forces and two moments, excluding torsion about the pin axis—transmitted through a central 2 mm stainless steel pin. While prior work has employed plated through-holes for capacitive sensing~\cite{Kim2020SixAxis}, our design differs by vertically stacking the electrodes and routing the applied load directly through the central pin, resulting in a significantly reduced form factor.

\subsection{Simulation}

We validated the sensor design using finite element analysis (FEA) in COMSOL Multiphysics. The model consists of a 3 mm diameter plated through-hole with a concentrically positioned M2 pin (2 mm diameter) surrounded by silicone elastomer. 
We assigned a dielectric constant of 2.8 to the silicone layer (from the COMSOL material library), and set the PCB thickness to 1.6 mm. The pin was electrically grounded while each electrode segment was connected to a 1 V terminal to compute Maxwell self-capacitance.

\begin{figure}[tbp!]
	\centering
	\begin{subfigure}[h]{0.99\linewidth}
	\centering
	\includegraphics[page=1, width=\linewidth]{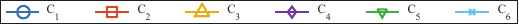}
	\end{subfigure}

	\centering
	  
        \begin{subfigure}[h]{0.32\linewidth}
    	\centering    	
        \includegraphics[page=2, width=\linewidth]{Figures/fig3_comsol.pdf}
    	\caption{Lateral displacement}
    	\label{fig:fea_force_vs_pressure}
        \end{subfigure}
        \vspace{5mm}
        \hfill
        \begin{subfigure}[h]{0.32\linewidth}
    	\centering    	
         \includegraphics[page=3, width=\linewidth]{Figures/fig3_comsol.pdf}
    	\caption{Vertical displacement}
    	\label{fig:fea_force_vs_pressure}
        \end{subfigure}
        \begin{subfigure}[h]{0.32\linewidth}
    	\centering    	
        \includegraphics[page=4, width=\linewidth]{Figures/fig3_comsol.pdf}
    	\caption{Rotational displacement}
    	\label{fig:fea_force_vs_pressure}
        \end{subfigure}
    \caption{Simulated capacitance changes of the six sensing channels in response to central pin displacement: (a) lateral displacement along the $x$-axis, (b) vertical displacement along the $z$-axis, and (c) rotational displacement about $\theta$.}
    \vspace{-3mm}
    \label{fig:COMSOL}
\end{figure}

We performed electrostatic simulations under three fundamental deformation modes: lateral displacement in the x-direction, vertical displacement in the z-direction, and pure rotation about the y-axis, as illustrated in \cref{fig:COMSOL}. For representative deformations (x = 0.3 mm, z = 0.3 mm, $\theta$ = 3.5°), each mode produced distinct capacitance distributions across the six electrode channels. Peak capacitance changes from the nominal state ranged from 0.08 to 0.27 pF—well within the measurable range of commercial capacitive-to-digital converters such as the AD7147. These results confirm that different force and torque combinations produce unique six-channel capacitance signatures suitable for calibration-based force/torque estimation.

To test design sensitivity, we simulated pin diameters of 1.5, 2.0, and 2.5 mm with fixed 3 mm hole diameter and x = 0.2 mm displacement. Peak capacitance changes increased over 10× from 0.06 pF (1.5 mm) to 0.71 pF (2.5 mm) as the annular gap decreased. However, the 2.5 mm pin's 0.25 mm gap complicates uniform silicone filling and bonding. We selected the 2.0 mm diameter (M2 pin) to balance sensitivity and manufacturability.

\subsection{Sensor Fabrication}

The sensor consists of three stacked PCBs. The top and bottom PCBs form the sensing electrodes, while the middle PCB integrates a capacitance-to-digital converter (CDC chip, AD7147). The CDC supports a 16 pF input range with 16-bit resolution, which sufficiently covers the capacitance variation predicted by simulation.

\begin{figure}[t]
    \centering

    \begin{subfigure}[t]{0.48\columnwidth}
        \centering
        \includegraphics[width=\linewidth]{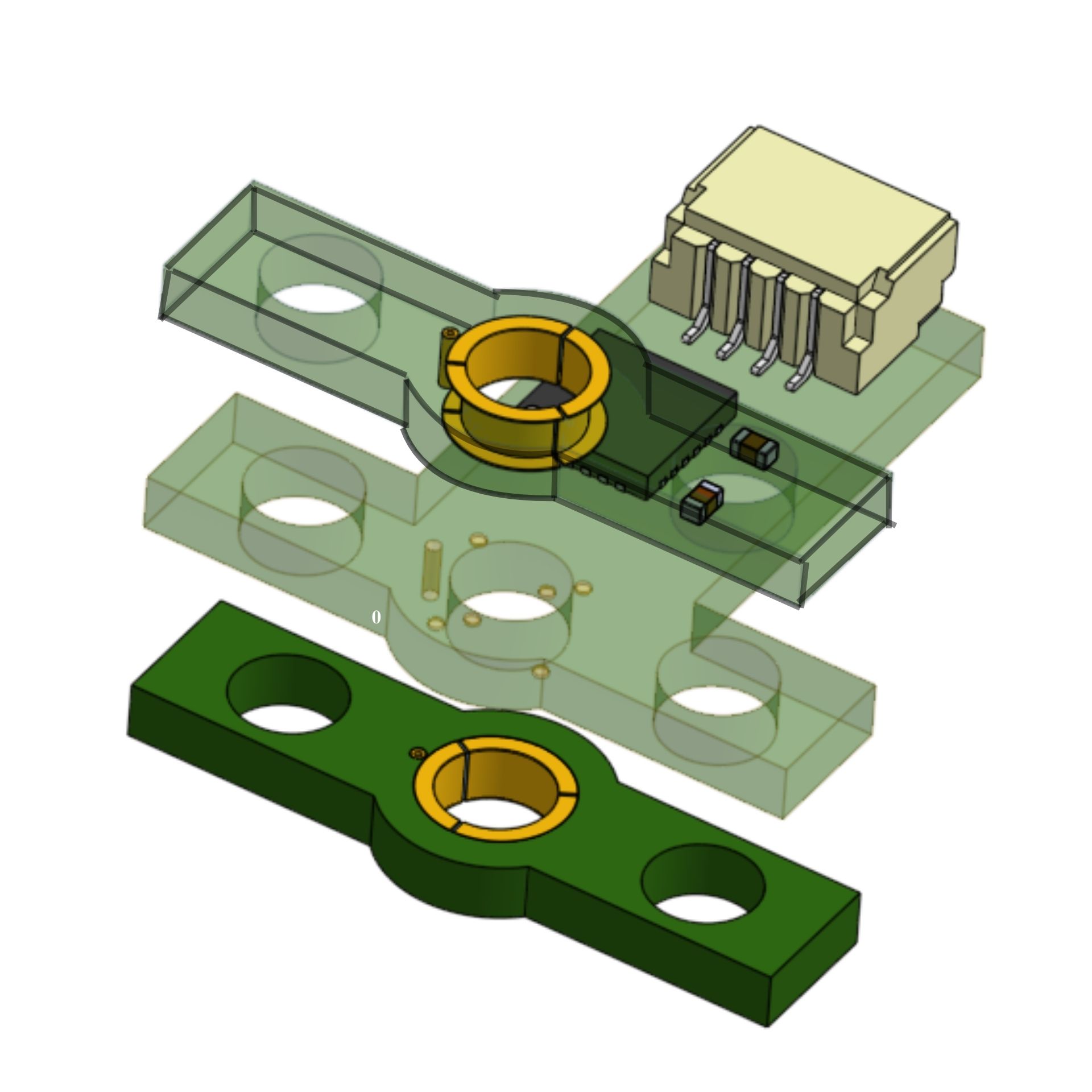}
        \caption{ Exploded view of sensor layers}
        \label{fig:pcbStack}
    \end{subfigure}
    \hfill
    \begin{subfigure}[t]{0.48\columnwidth}
        \centering
        \includegraphics[width=\linewidth]{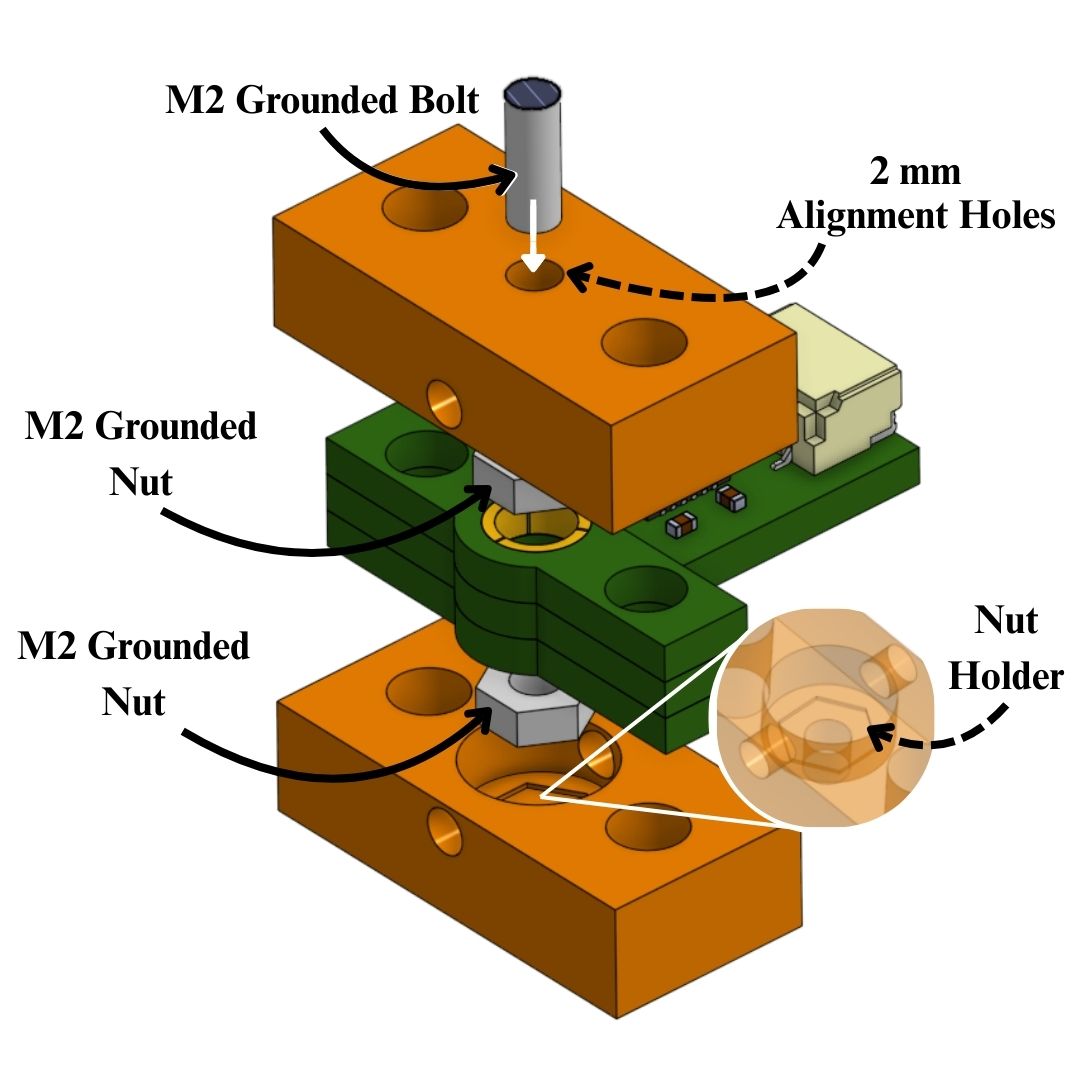}
        \caption{Silicone casting mold alignment}
        \label{fig:sensorAlign}
    \end{subfigure}

    \begin{subfigure}[t]{0.48\columnwidth}
        \centering
        \includegraphics[width=\linewidth]{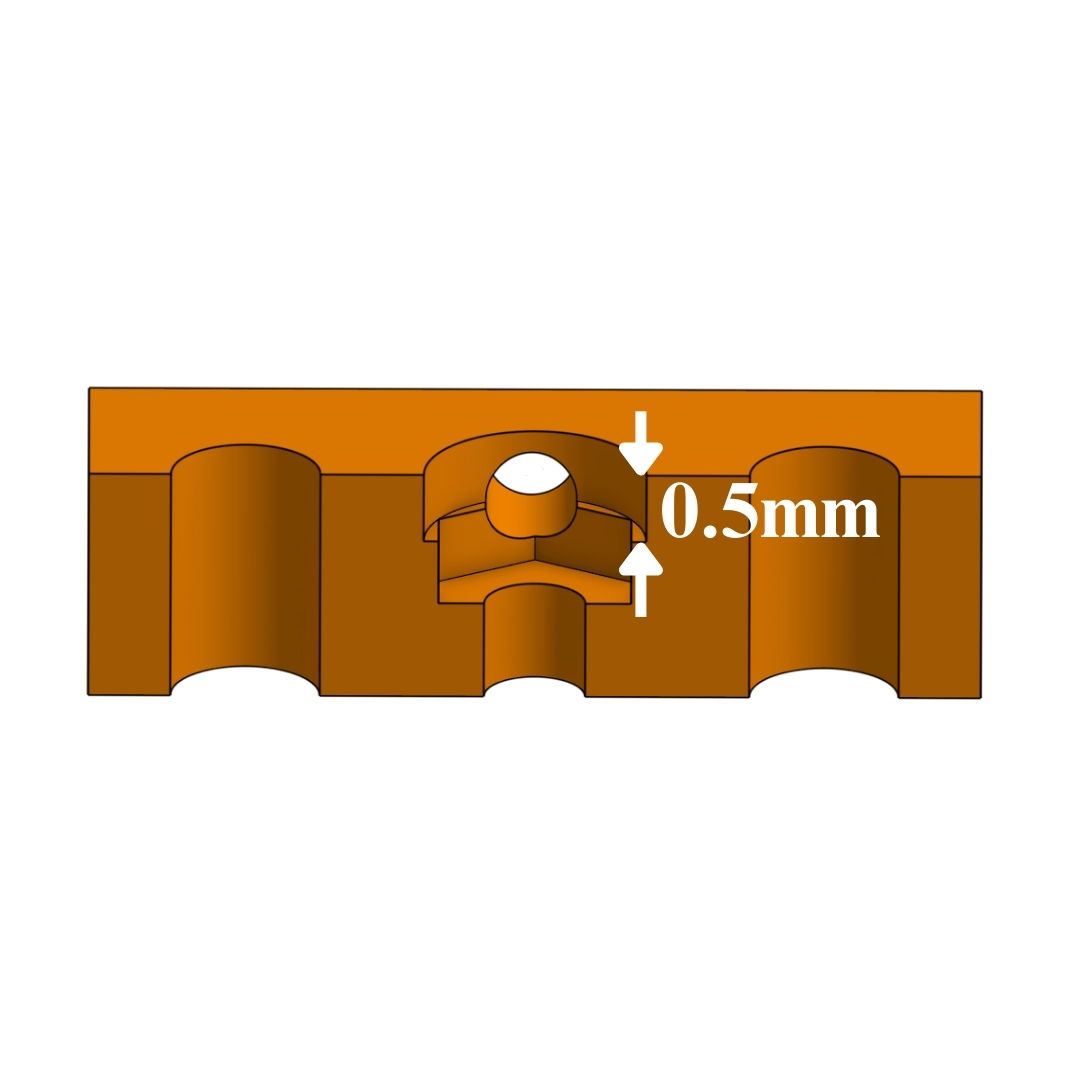}
        \caption{ Cast mold cross-section}
        \label{fig:CastMold}
    \end{subfigure}
    \hfill
    \begin{subfigure}[t]{0.48\columnwidth}
        \centering
        \includegraphics[width=\linewidth]{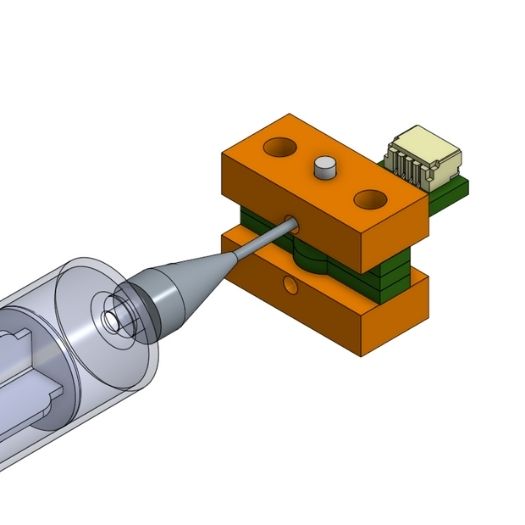}
        \caption{ Silicone injection}
        \label{fig:Syringed}
    \end{subfigure}


    \begin{subfigure}[t]{0.31\columnwidth}
        \centering
        \includegraphics[width=\linewidth]{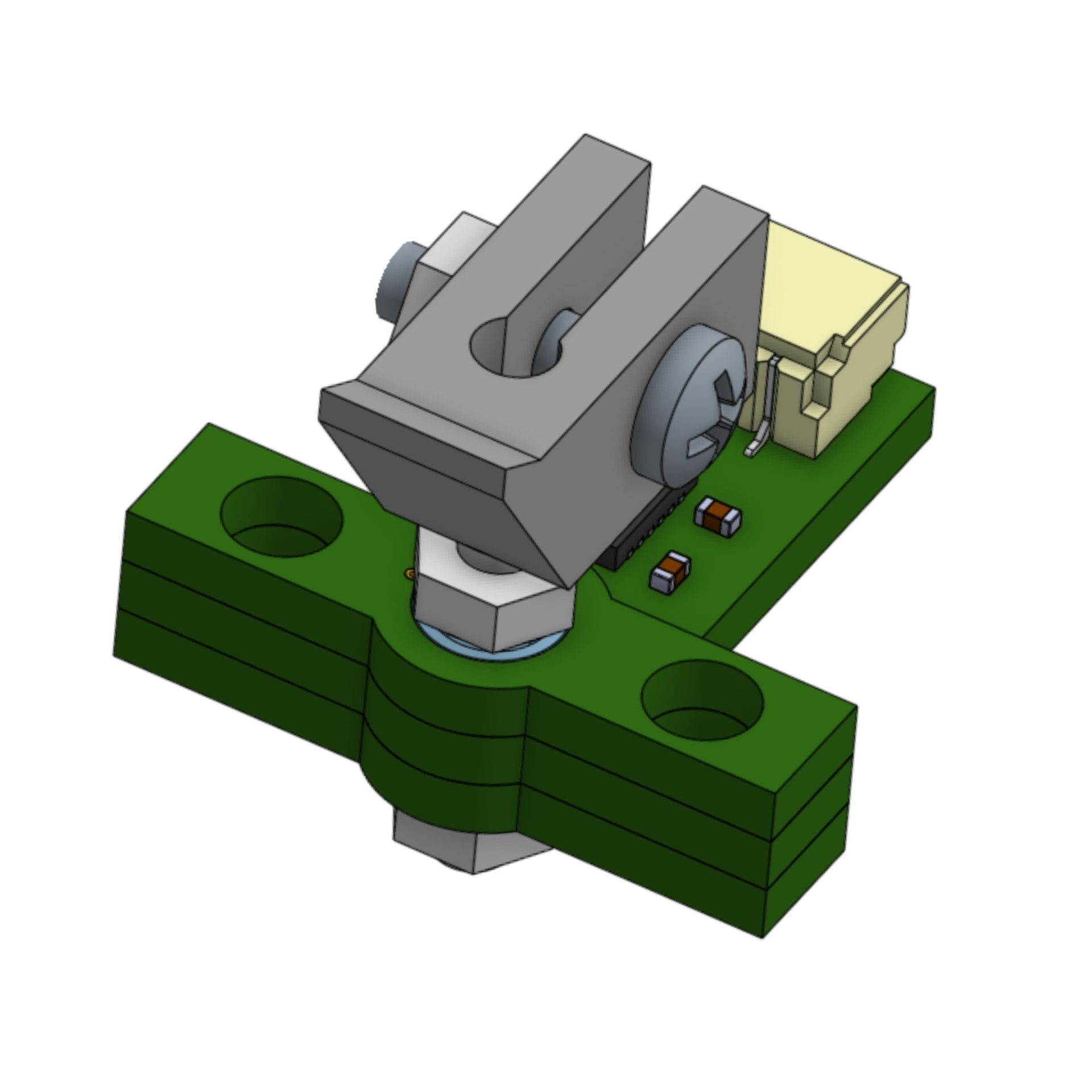}
        \caption{Sensor with Tip}
        \label{fig:CADSensorModel}
    \end{subfigure}
    \hfill
    \begin{subfigure}[t]{0.31\columnwidth}
        \centering
        \includegraphics[width=\linewidth]{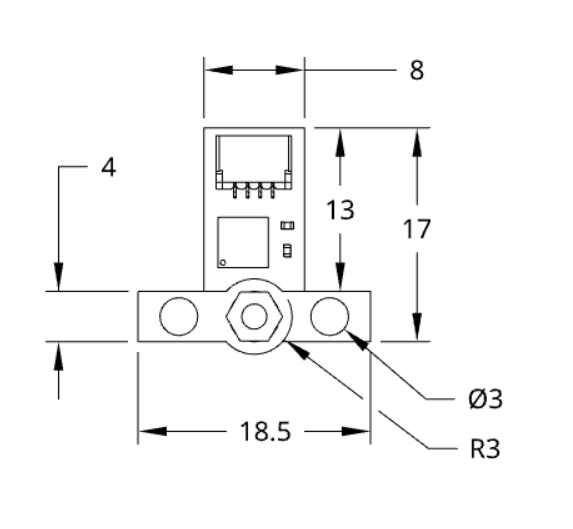}
        \caption{Top view}
        \label{fig:TopViewDimensions}
    \end{subfigure}
    \hfill
    \begin{subfigure}[t]{0.31\columnwidth}
        \centering
        \includegraphics[width=\linewidth]{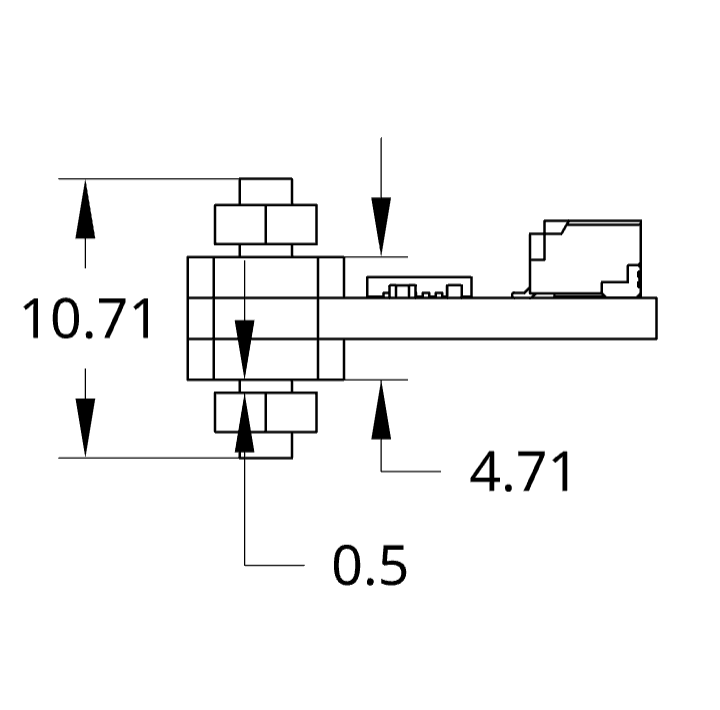}
        \caption{Side view}
        \label{fig:SideViewDimensions}
    \end{subfigure}
    \vspace{1mm}
    \caption{Sensor manufacturing workflow: electrode stacking, silicone casting, and final CAD geometry. All dimensions are in millimeters.}
    \label{fig:SolderingProcess1}
\end{figure}

As shown in \cref{fig:pcbStack}, each PCB contains three 3 mm through-holes. The central plated through-hole on the top and bottom PCBs serves as the sensing electrode, which is segmented into three electrically isolated sections using a 0.2 mm wire saw. The two side through-holes are used for alignment and mounting during assembly and can be removed in future designs to further reduce the sensor footprint.

During assembly, each segmented electrode on the top and bottom PCBs is electrically connected to the CDC on the middle PCB by forming solder balls on the middle PCB and reflowing them through the corresponding solder joint vias. The three aligned PCBs are bonded using cyanoacrylate adhesive to create a rigid stacked structure.

A 2 mm-diameter stainless steel dowel pin, shown in \cref{fig:sensorAlign}, serves as the grounded central electrode. The pin is M2-threaded at both ends to accommodate constraining nuts, while the central sensing region remains unthreaded to provide a smooth surface for silicone adhesion. A custom 3D-printed casting mold (\cref{fig:CastMold}) is used to position the pin and nuts and to define the silicone geometry. The mold includes inlet and outlet ports for silicone injection and constrains the nut locations to maintain a uniform 0.5 mm silicone gap between the nuts and the PCB surfaces.

Silicone adhesive (Sil-Poxy, Smooth-On) is injected using a syringe (\cref{fig:Syringed}) until material exits the outlet port, indicating complete filling of the sensing region. After curing, the mold is removed, leaving a compliant silicone layer encapsulating the sensing region. A compact tweezer tip (\cref{fig:CADSensorModel}) is subsequently fabricated via 3D printing and mechanically clamped onto the exposed end of the central pin.

\begin{figure*}[tp!]
    \centering
    \renewcommand{\thesubfigure}{\alph{subfigure}}

    \begin{subfigure}[t]{0.32\textwidth}
        \centering
        \includegraphics[width=\linewidth]{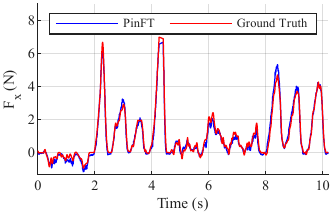}
        \caption{Est./True F$_x$}
        \label{fig:SilFittedFx}
    \end{subfigure}
    \hfill
    \begin{subfigure}[t]{0.32\textwidth}
        \centering
        \includegraphics[width=\linewidth]{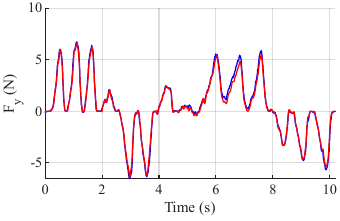}
        \caption{Est./True F$_y$}
        \label{fig:SilFittedFy}
    \end{subfigure}
    \hfill
    \begin{subfigure}[t]{0.32\textwidth}
        \centering
        \includegraphics[width=\linewidth]{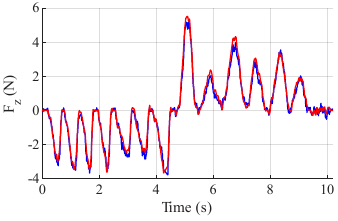}
        \caption{Est./True F$_z$}
        \label{fig:SilFittedFz}
    \end{subfigure}

    \vspace{0.6em}

    \begin{subfigure}[t]{0.32\textwidth}
        \centering
        \includegraphics[width=\linewidth]{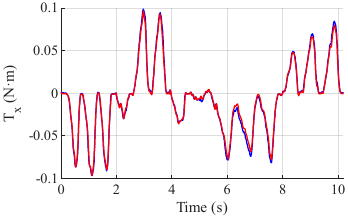}
        \caption{Est./True T$_x$}
        \label{fig:SilFittedTx}
    \end{subfigure}
    \hfill
    \begin{subfigure}[t]{0.32\textwidth}
        \centering
        \includegraphics[width=\linewidth]{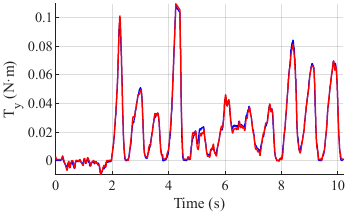}
        \caption{Est./True T$_y$}
        \label{fig:SilFittedTy}
    \end{subfigure}
    \hfill
    \begin{subfigure}[t]{0.32\textwidth}
        \centering
        \includegraphics[width=\linewidth]{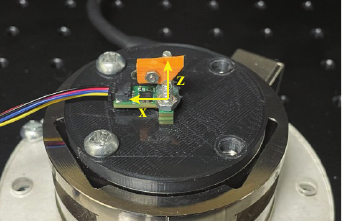}
        \caption{Calibration Setup}
        \label{fig:ATIMounted}
    \end{subfigure}

    \vspace{1mm}
    \caption{(a–e) Comparison between PinFT estimated and ground-truth forces (N) and torques (N·m) over time. (f) Final assembled PinFT sensor mounted on the calibration platform.
    }
    \label{fig:SilFittedFull}
\end{figure*}

\section{Sensor Characterization}
\label{chap:SensorPerformance}

Using the fabricated sensor, we calibrated it against force/torque inputs applied on a tweezer tip. As shown in \cref{fig:ATIMounted}, the PinFT with tweezertip is rigidly mounted to the ATI Gamma F/T sensor via a 3D-printed interface. Ground-truth forces and torques were measured using an ATI Gamma F/T sensor, sampled at 200~Hz through a LabJack T7 DAQ. The fabricated PinFT sensor was sampled via an ESP32 microcontroller communicating with the CDC at 220~Hz. Synchronization between the two systems was achieved using a digital sync pin from the ESP32 to the LabJack DAQ. The resulting signals from ATI Gamma F/T were linearly interpolated in MATLAB to match sampling rates. We manually applied forces and torques to the tweezer tip block to span the intended dynamic range, totaling 190,000 samples, or 15.7 minutes of equivalent data to calibrate (interpolated CDC to 200Hz to align data). 

For the calibration model, we seek a higher-order multivariable mapping, similar to \cite{choi2025coinft} as:
\begin{equation}
    \hat{\mathbf{y}} = \mathbf{M}\,\boldsymbol{\phi}(\mathbf{C}),
    \label{eq:CalibrationModel}
\end{equation}
where $\mathbf{C} \in \mathbb{R}^{6}$ is a vector of capacitance count change from nominal value of 6 channel, $\boldsymbol{\phi}(\cdot)$ builds a feature vector from $\mathbf{c}$, and $\mathbf{y} \in \mathbb{R}^{6}$ is an estimate of \{$F_x,F_y,F_z,T_x,T_y$\}, axis denoted in \cref{fig:CombinedElectrodeViews}. We estimate $\mathbf{M}$ by a least square fit. For $\boldsymbol{\phi}(\mathbf{c})$, we chose 1st and 2nd order polynomial of each capacitance (similar to \cite{choi2025coinft}) and additional cross product up to quadruple accounting for highly coupled capacitance measures:
\vspace{-3mm}
\begin{multline}
\boldsymbol{\phi}(\mathbf{C}) = \big[\, C_i,\; C_i^2,\; C_i C_j,\; C_i C_j C_k,\; C_i C_j C_k C_l \,\big]^\top \in\mathbb{R}^{77}\\
i,j,k,l \in \{1,\ldots,6\} \quad \text{all distinct}
\label{eq:MatFTFinal}
\end{multline}

\begin{table}[t]
  \centering
  \footnotesize
  \renewcommand{\arraystretch}{0.9}
  \setlength{\tabcolsep}{3pt}

  \begin{tabular}{p{1.5cm}ccccc}
    \toprule
        & \textbf{F$_x$} 
        & \textbf{F$_y$}
        & \textbf{F$_z$}
        & \textbf{T$_x$}
        & \textbf{T$_y$}
        \\ 
    \midrule
    \textbf{F/T Range}
        & $\pm 7$ N
        & $\pm 7$ N
        & $\pm 9$ N
        & $\pm 0.1$ Nm
        & $\pm 0.1$ Nm
        \\ 

    \textbf{MAE}
        & 0.22 N 
        & 0.20 N
        & 0.23 N
        & 2.6 mNm
        & 2.0 mNm
        \\ 

    \textbf{95pct$|\text{Err}|$}
        & 0.61 N
        & 0.55 N
        & 0.61 N
        & 7.2 mNm
        & 5.9 mNm
        \\ 

    \textbf{$\text{R}^2$}
        & 0.97
        & 0.99
        & 0.97
        & 0.99
        & 0.99
        \\

    \textbf{Noise (1$\sigma$)}
        & 16 mN
        & 8.8 mN
        & 50 mN
        & 0.13 mNm
        & 0.19 mNm
        \\ 
        \bottomrule

  \end{tabular}
  \vspace{1.5mm}
  \caption{PinFT calibrated performance specifications reporting calibrated range, mean absolute error (MAE), 95th-percentile absolute error ($95\text{pct}|\text{Err}|$), coefficient of determination ($R^2$), and noise ($1\sigma$) for each sensing axis. All metrics are computed on test data excluded from calibration training.
}
  \label{tab:ft_methods_clean}
\end{table}

\subsection{Calibration Results}

As shown in \cref{fig:SilFittedFull} and summarized in \cref{tab:ft_methods_clean}, the calibrated 5-DOF PinFT sensor demonstrates close agreement with the ground-truth force/torque measurements. Across the held-out test dataset, the mean absolute error (MAE) is approximately 0.23~N for forces and 2.5~mN$\cdot$m for torques, with coefficients of determination ($R^2$) exceeding 0.97 for all axes. These results indicate that the proposed calibration effectively captures the dominant multi-axis force--torque relationships within the tested operating regime.

The calibration range is limited to forces below 9~N and torques below 0.1~N$\cdot$m, primarily due to the mechanical robustness of the bonded structure comprising the Sil-Poxy elastomer, pins, and stacked PCBs. While this range is narrower than that of commercial force/torque sensors, it is sufficient for the delicate manipulation tasks targeted in this work, as demonstrated in the following sections. Within a comparable force range, the achieved force and torque errors are on par with those reported for CoinFT~\cite{choi2025coinft}, while PinFT offers a substantially smaller sensor footprint.

The 95th-percentile absolute error (95pct$\lvert\mathrm{Err}\rvert$) is slightly higher than the MAE, with force errors remaining below approximately 0.6~N across all axes. This corresponds to 7.8\% of full scale and reflects worst-case performance rather than average accuracy. Given the low-cost construction and compact form factor of the proposed sensor, this level of accuracy is acceptable for applications that require coarse force estimation combined with continuous contact monitoring, rather than metrology-grade force measurement. In addition, the measured noise levels are low relative to the operating force range, supporting reliable detection of contact onset and changes in interaction forces. The $F_z$ axis exhibits higher noise due to its lower sensitivity resulting from the sensor design.

\section{Robotic Tweezer Application}
\label{chap:RobotDemo}

To evaluate the performance of PinFT in practical robotic tasks, a 3D-printed tweezer integrated with PinFT was mounted on a parallel-jaw gripper to manipulate selected objects while measuring tool-tip force/torque. The experiments demonstrate characteristic force/torque signatures corresponding to stable grasping, slip, and tearing events.

\subsection{Experiments}
\begin{figure}[t]
    \centering

    \begin{subfigure}[t]{0.48\columnwidth}
        \centering
        \includegraphics[width=\linewidth]{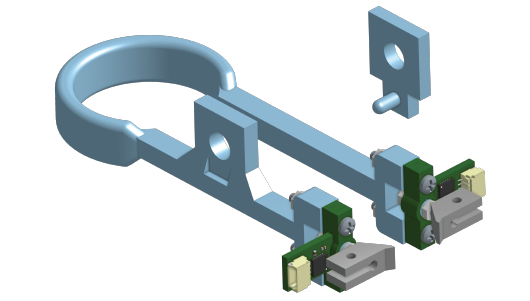}
        \caption{ Tweezer assembly}
        \label{fig:CompleteTweezer}
    \end{subfigure}
    \hfill
    \begin{subfigure}[t]{0.48\columnwidth}
        \centering
        \includegraphics[width=\linewidth]{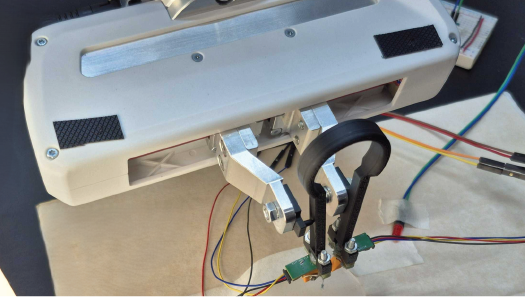}
        \caption{ Tweezer on parallel-jaw gripper}
        \label{fig:RobotTweezer}
    \end{subfigure}

    \vspace{0.4em}


    \caption{Tweezer design integrated with PinFT and attachment to a robotic parallel-jaw gripper.}
    \label{fig:TweezerDesignCombined}
\end{figure}

We 3D-printed a PLA tweezer, as shown in \cref{fig:TweezerDesignCombined}, that integrates PinFT sensors on both tip sides. The body incorporates an elastic loop for passive reopening, and for robotic actuation, one handle is rigidly bolted to a parallel-jaw gripper finger while a pusher mounted on the opposing finger closes the tweezer.

\begin{figure}[t]
    \centering
    \begin{subfigure}[t]{0.31\columnwidth}
        \centering
        \includegraphics[width=\linewidth]{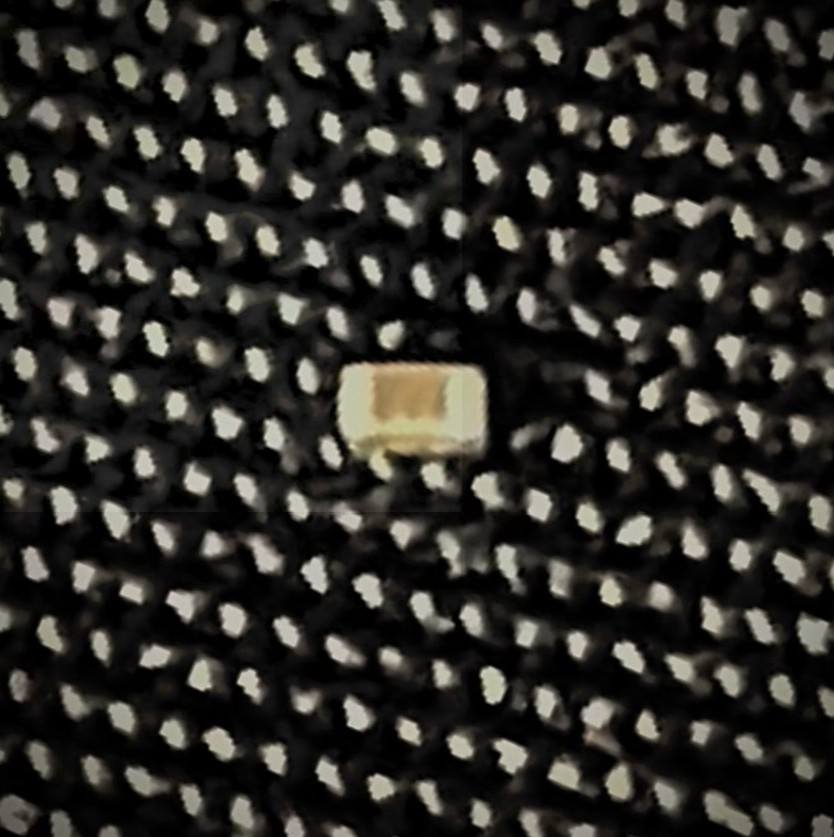}
        \caption{SMD Capacitor}
        \label{fig:ObjectSMD}
    \end{subfigure}
    \hfill
    \begin{subfigure}[t]{0.31\columnwidth}
        \centering
        \includegraphics[width=\linewidth]{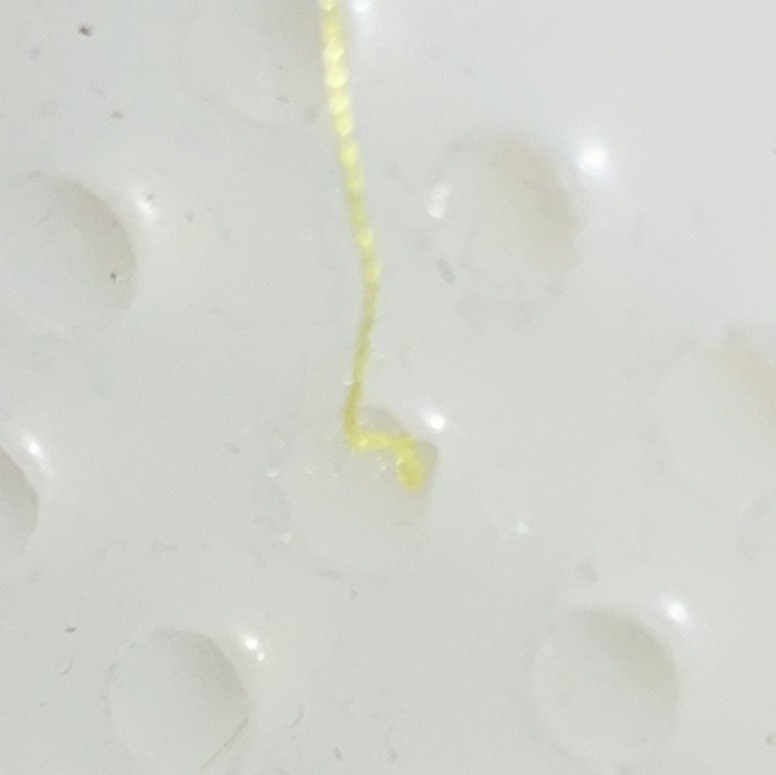}
        \caption{Simulated Hair}
        \label{fig:ObjectHairSilicone}
    \end{subfigure}
    \hfill
    \begin{subfigure}[t]{0.31\columnwidth}
        \centering
        \includegraphics[width=\linewidth]{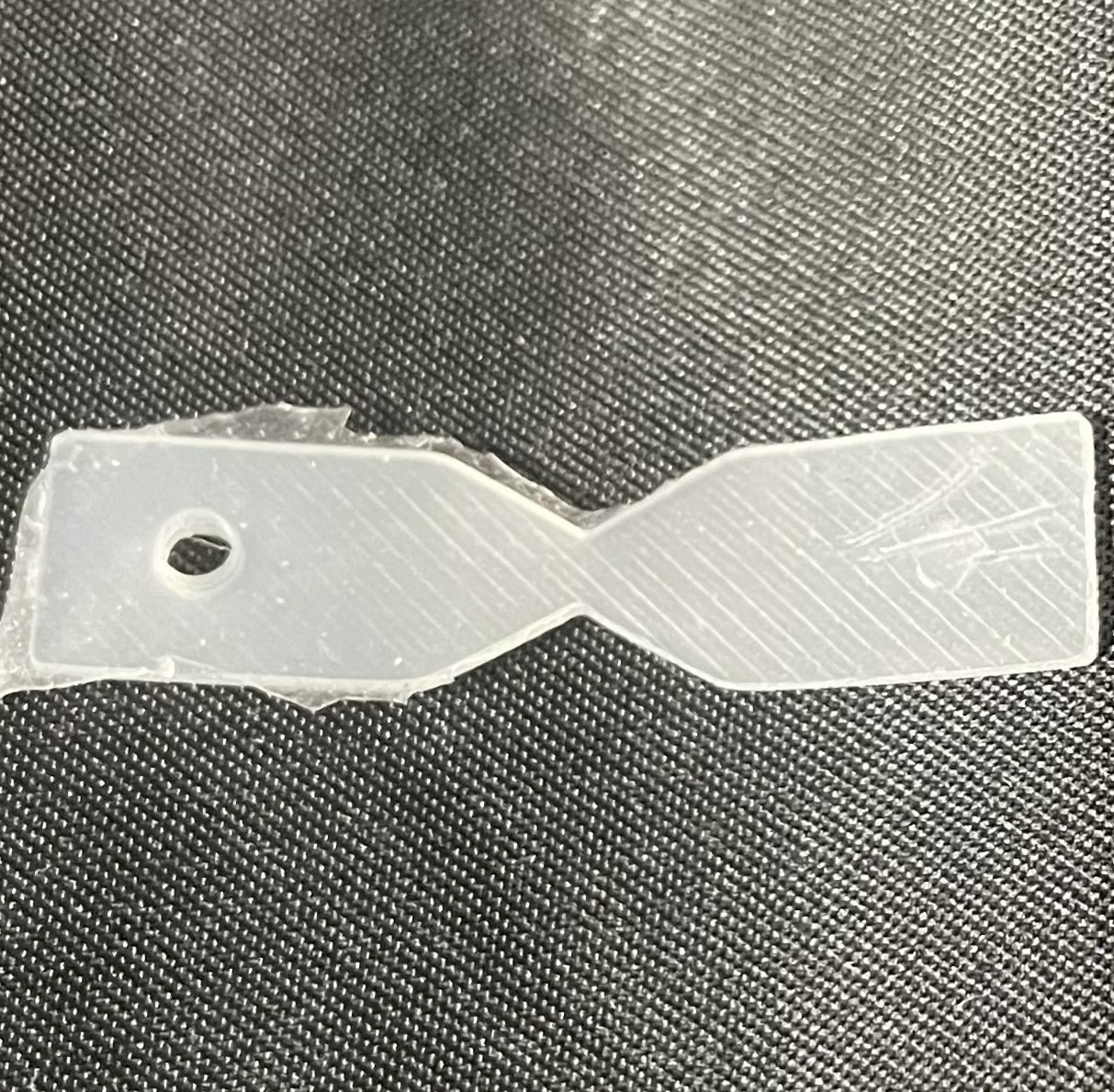}
        \caption{Silicone Test Piece}
        \label{fig:ObjectSiliconeTestPiece}
    \end{subfigure}
    \vspace{1mm}
    \caption{Test objects used in tweezing experiments: (a) SMD capacitor (L$\times$W$\times$H = 
0.4$\times$0.2$\times$0.2~mm), (b) simulated hair embedded in silicone (diameter = 1~mm), 
and (c) silicone test piece (L$\times$W$\times$H = 60$\times$20$\times$1~mm, Ecoflex-0030).}
    \label{fig:ObjectsTweezing}
\end{figure}

Testing tasks, shown in \cref{fig:ObjectsTweezing}, were selected to represent typical tweezer operations: picking up a small surface-mount capacitor, pulling a simulated hair embedded in a silicone block (string surrogate), and pulling/tearing a compliant silicone specimen. These tasks span delicate, fine-scale, and compliant interactions, enabling evaluation of the sensor’s ability to capture direct tool-tip forces under diverse contact conditions.

The PLA tweezer with integrated PinFT sensors was mounted on the parallel-jaw gripper of a Franka Emika Research 3 robot. As the focus of this work is force sensing rather than vision perception or motion planning, the arm was manually positioned to the desired engagement pose, after which the gripper closed to a user-defined width and executed lifting or pulling motions. Closing widths were selected from preliminary trials to produce both secure and borderline grasps, allowing comparison of successful and unsuccessful manipulation, while embedded force/torque data were collected throughout each trial.

\subsection{Results}

\subsubsection{Picking up a Small SMD Capacitor}
\begin{figure}[t]
    \centering


    \begin{subfigure}[t]{0.49\linewidth}
        \centering
        \includegraphics[width=\linewidth, page=1]{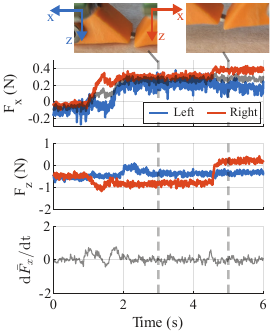}
        \caption{Capacitor Grasped}
        \label{fig:SMDSuccess}
    \end{subfigure}
    \hfill
    \begin{subfigure}[t]{0.49\linewidth}
        \centering
        \includegraphics[width=\linewidth, page=2]{Figures/figure8_SMD_copy.pdf}
        \caption{Capacitor Ejected}
        \label{fig:SMDFailure}
    \end{subfigure}

    \vspace{1mm}
    \caption{Tweezer grasping experiment on an SMD capacitor. The black line in $F_x$ represents the mean of the left and right grasp forces. Slip events are indicated by a sudden drop in grasp force, reflected as a negative peak in $d\bar{F}_x/dt$ at the moment of ejection. Dashed vertical lines denote key transition events.}
    \label{fig:SMDTest}
\end{figure}

In grasping small SMD capacitors, among the 5-DOF measurements, $F_x$ and $F_z$ exhibit the most task-relevant force profiles, as shown in \cref{fig:SMDTest}. When grasping such small objects on a table surface, the tweezer tips must make contact with the table to ensure proper engagement, resulting in negative $F_z$ forces during the approach phase. The individual tip sensors enable detection of which tip contacts the table first; in both cases shown in \cref{fig:SMDTest}, the right tip exerts greater force on the table than the left tip. This differential contact information can be exploited for tip contact force balancing control.

As the tweezer closes, $F_x$ increases, corresponding to the internal grasping force. However, for small object manipulation, grasp force magnitude alone is not a sufficient metric for success. Excessive grasp force can cause the object to slip or eject from between the tips, as illustrated in \cref{fig:SMDFailure}. When the object is lost, a sudden decrease in the mean grasp force (denoted $\bar{F_x}$) is observed.

To quantify this event, we computed $d\bar{F_x}/dt$ after smoothing the signal using a Savitzky--Golay filter (3rd order polynomial, window size 41 samples). The gradient of the filtered signal is shown in the bottom plot of \cref{fig:SMDTest}. Upon object ejection, a pronounced negative gradient is observed, consistent with the values summarized in \cref{tab:derivatives}. 

Importantly, this internal grasp-force change, as well as the balancing behavior in $F_z$ between the two tips, can only be detected by having force/torque sensing at each individual tweezer tip. Such individual and internal force interactions cannot be measured by an external wrist-mounted force/torque sensor, which captures only net interaction forces at the robot interface rather than internal grasp forces between the tips.

\begin{figure}[t]
    \centering

    \begin{subfigure}[t]{0.49\linewidth}
        \centering
        \includegraphics[width=\linewidth, page=1]{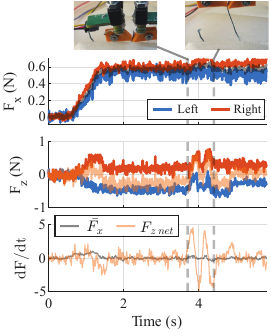}
        \caption{Simulated Hair Pulled Out}
        \label{fig:HairSuccess}
    \end{subfigure}
    \hfill
    \begin{subfigure}[t]{0.49\linewidth}
        \centering
        \includegraphics[width=\linewidth, page=2]{Figures/figure9_wire_copy.pdf}
        \caption{Simulated Hair Slip}
        \label{fig:HairFailure}
    \end{subfigure}

    \vspace{1mm}
    \caption{Tweezer grasping and pull experiment on simulated hair showing (a) successful pullout 
and (b) slip during pull. In the successful case, $dF/dt$ fluctuations appear 
only in the pull force ($F_{z,\text{net}}$) while the grasp force ($\bar{F}_x$) 
remains stable. In the slip case, negative peaks are observed in both the pull force and grasp force, indicating loss of grip.  }
    \label{fig:HairTest}
\end{figure}

\subsubsection{Pulling Simulated Hair}
\label{sec:hairResult}
Without prior knowledge of the required pull force or internal grasp force, we monitor pull success by observing the two-axis force components $F_x$ and $F_z$, as shown in \cref{fig:HairTest}. 

When the hair is successfully pulled out, the internal grasping force maintains the hair within the tweezer while the vertical force is applied for extraction. This is evident in the bottom panel of \cref{fig:HairSuccess}, where during the pulling phase (between the two dashed vertical lines), $dF_{z,\text{net}}/dt$ exhibits noticeable variations due to the elimination of pulling resistance from the silicone substrate. In contrast, $d\bar{F}_x/dt$, representing the mean grasp force, remains relatively constant throughout the pulling motion. 

Conversely, when the hair pull fails, the gripper exhibits a noticeable negative $dF_{z,\text{net}}/dt$ accompanied by negative $d\bar{F}_x/dt$. This concurrent decrease occurs because the pulling force on the hair ceases as the gripper loses its grip through slippage. This is also shown in the data summary in \cref{tab:derivatives}. 

These results demonstrate that multi-axis tip-level sensing enables clear differentiation between object-environment interactions (i.e., hair pull-out) and object-tweezer events (i.e., tweezer slips on the hair), providing critical information for pullout success detection.

\begin{figure}[t]
    \centering
       \begin{subfigure}[t]{0.49\linewidth}
        \centering
        \includegraphics[width=\linewidth, page=1]{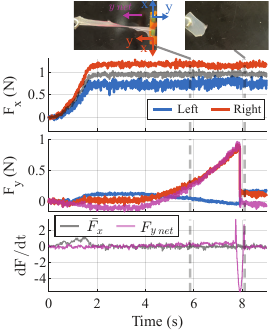}
        \caption{ Silicone Pullout and Tear}
        \label{fig:SiliconeSuccess}
    \end{subfigure}
    \hfill
    \begin{subfigure}[t]{0.49\linewidth}
        \centering
        \includegraphics[width=\linewidth, page=2]{Figures/figure10_silicone_copy.pdf}
        \caption{Silicone Slip}
        \label{fig:SiliconeFailure}
    \end{subfigure}

    \vspace{1mm}
    \caption{Tweezer pull and tearing experiment on a silicone test piece showing 
(a) successful pullout and tearing and (b) slip during pull. In the successful 
case, $dF/dt$ fluctuations appear only in the pull force ($F_{y,\text{net}}$) 
while the grasp force ($\bar{F}_x$) remains stable. In the slip case, negative 
peaks are observed in both the pull force and grasp force; an attenuation in the rising slope of $F_{y,\text{net}}$ prior to the slip 
event indicates sliding of the piece before full loss of grip. }
    \label{fig:SiliconeTest}
\end{figure}

\subsubsection{Tearing Silicone Specimen}
In this experiment, the silicone specimen is grasped and pulled along the y-axis until the silicone block either tears or slips. The robot pull motion was commanded with identical control parameters in all trials. As shown in \cref{fig:SiliconeTest} and \cref{tab:derivatives}, successful tear and slip failure cases exhibit similar patterns to those observed in \cref{sec:hairResult}. Successful tears are characterized by a drop in $dF_{y,\text{net}}/dt$ while maintaining $\bar{F}_x$, whereas failure cases show a concurrent drop in both $dF_{y,\text{net}}/dt$ and $d\bar{F}_x/dt$.

An additional distinguishing characteristic is the slope of $F_{y,\text{net}}$ prior to the sudden drop. In successful tear cases, $F_{y,\text{net}}$ exhibits a consistent ramp-up followed by an abrupt drop at rupture, while in failure cases, the slope attenuates before failure occurs. This slope change is attributed to incipient slip rather than variations in grasp force, as \( F_x \) (the grasp force) remains steady or slightly increases during the event. When handling deformable compliant materials, such slope attenuation can serve as an early indicator of incipient slip, enabling adaptive control to increase grasp force preemptively and prevent failure. This capability is particularly valuable for handling delicate specimens where excessive grasp force is undesirable—tip-level multi-axial sensing may provide the necessary feedback to appropriately adjust the grasp force.

\begin{table}[tbp!]
\centering
\scriptsize 
\setlength{\tabcolsep}{3pt}
\newcolumntype{C}[1]{>{\centering\arraybackslash}p{#1}}
\begin{tabular}{p{0.8cm}*{4}{C{1.6cm}}}
\toprule
& \multicolumn{2}{c}{Success} & \multicolumn{2}{c}{Failure} \\
\cmidrule(lr){2-3} \cmidrule(lr){4-5}
& Min($d\bar{F}_x/dt$) & Min($dF_{\text{net}}/dt$) & Min($d\bar{F}_x/dt$) & Min($dF_{\text{net}}/dt$) \\
\midrule
\textbf{SMD}      & -0.68 $\pm$ 0.08 & --                & -1.44 $\pm$ 0.32 & --                \\
\textbf{Hair}     & -0.44 $\pm$ 0.05 & -3.44 $\pm$ 1.29  & -0.85 $\pm$ 0.23 & -2.89 $\pm$ 0.45  \\
\textbf{Silicone} & -0.62 $\pm$ 0.21 & -4.24 $\pm$ 1.48  & -1.03 $\pm$ 0.14 & -4.38 $\pm$ 0.15  \\
\bottomrule
\vspace{.5mm}
\end{tabular}
\caption{Negative peak $dF/dt$ values (mean $\pm$ std) over three trials for each object 
and outcome (Success/Failure). $\bar{F}_x$ denotes the grasp force, and 
$F_{\text{net}}$ denotes the net pull-direction force: $F_z$ for Hair and $F_y$ 
for Silicone.}
\label{tab:derivatives}
\end{table}

\section{Conclusion}

We presented \textit{PinFT}, a compact five-axis capacitive force/torque sensor designed for direct tool-tip integration in tweezer-like tools. The sensor employs a three-PCB stack with segmented plated through-hole electrodes and a silicone elastomer dielectric, enabling five-degree-of-freedom force and torque sensing transmitted through a central M2 pin. Finite-element analysis guided the structural design, and calibration of the fabricated prototype yielded mean absolute errors of approximately 0.23~N for forces and 2.5~mN$\cdot$m for torques, with $R^2$ exceeding 0.97 across all measured axes.

When integrated into a robotic tweezer mounted on a parallel-jaw gripper, PinFT captured distinctive multi-axis force signatures across three representative manipulation tasks: grasping a small SMD capacitor, pulling a simulated hair from a silicone substrate, and tearing a compliant silicone specimen. In all cases, tip-level sensing enabled reliable discrimination between successful manipulation and failure modes such as slip and object ejection. Gradient-based features derived from internal grasp force and net interaction force proved particularly effective for detecting incipient failure, highlighting the importance of distal, per-tip sensing for fine manipulation.

The current design is limited to forces below 9~N and torques below 0.1~N$\cdot$m, primarily constrained by the mechanical robustness of the bonded silicone–PCB structure. Extending the force range would require stiffer dielectric materials or structural reinforcement, potentially reducing sensitivity. Future work will focus on real-time closed-loop force control using PinFT measurements to enable adaptive, failure-aware manipulation of delicate objects. In addition, the tool-level sensing capability may support learning-from-demonstration paradigms, where human manipulation data collected through the instrumented tweezer can inform robot policies for delicate handling tasks.

\bibliographystyle{IEEEtran}

\balance  

\bibliography{Biblio}   
\vspace{\baselineskip}


\end{document}